\documentclass[letterpaper]{article} % DO NOT CHANGE THIS
\usepackage{aaai24}  % DO NOT CHANGE THIS
\usepackage{times}  % DO NOT CHANGE THIS
\usepackage{helvet}  % DO NOT CHANGE THIS
\usepackage{courier}  % DO NOT CHANGE THIS
\usepackage[hyphens]{url}  % DO NOT CHANGE THIS
\usepackage{graphicx} % DO NOT CHANGE THIS
\usepackage{natbib}  % DO NOT CHANGE THIS AND DO NOT ADD ANY OPTIONS TO IT
\usepackage{caption} % DO NOT CHANGE THIS AND DO NOT ADD ANY OPTIONS TO IT
\usepackage{algorithm}
\usepackage{algorithmic}
\usepackage{amsmath}
\usepackage{amsthm}
\usepackage{newfloat}
\usepackage{listings}
\DeclareCaptionStyle{ruled}{labelfont=normalfont,labelsep=colon,strut=off} % DO NOT CHANGE THIS
\floatstyle{ruled}
\newfloat{listing}{tb}{lst}{}
\floatname{listing}{Listing}
\usepackage{enumitem}
\setlist[itemize]{left=2em, labelsep=0.5em}

\usepackage{booktabs}

\title{A Human-in-the-Loop Fairness-Aware Model Selection Framework \\ for Complex Fairness Objective Landscapes}
\author {
    Jake Robertson\textsuperscript{\rm 1, \rm 2},
    Thorsten Schmidt\textsuperscript{\rm 1},
    Frank Hutter\textsuperscript{\rm 1, \rm 3},
    Noor Awad\textsuperscript{\rm 1}
}
\affiliations {
    \textsuperscript{\rm 1}University of Freiburg, \\
    \textsuperscript{\rm 2}Zuse School ELIZA, \\
    \textsuperscript{\rm 3}ELLIS Institute Tübingen\\
    robertsj@cs.uni-freiburg.de, thorsten.schmidt@stochastik.uni-freiburg.de, \\ 
    fh@cs.uni-freiburg.de, awad@cs.uni-freiburg.de
}
\begin{document}

\maketitle

\newtheorem{definition}{Definition}[section]

\begin{figure*}[h!]
    \centering
    \includegraphics[width=\textwidth]{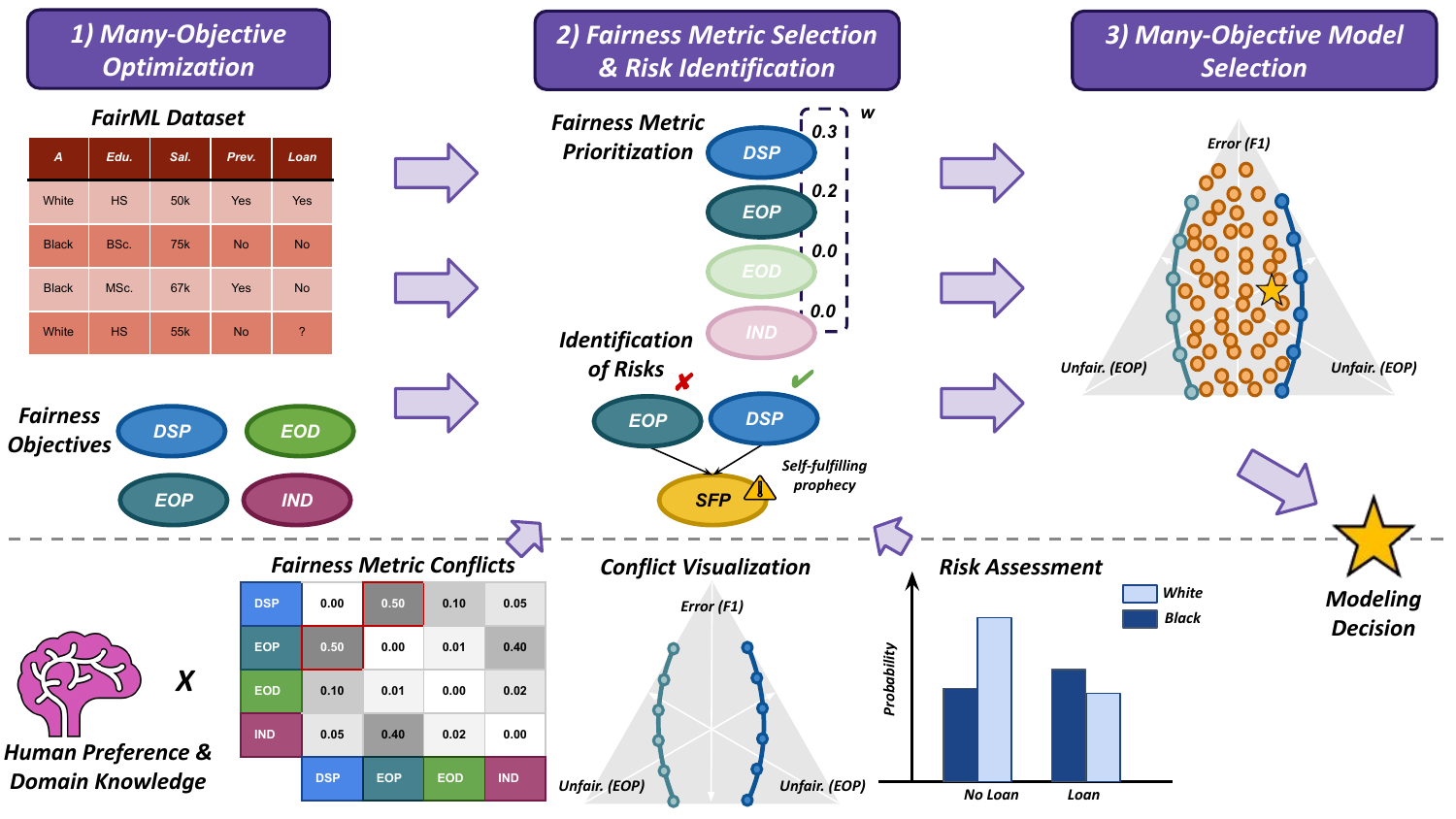}
    \caption{\textit{ManyFairHPO}: We introduce ManyFairHPO, a human-in-the-loop, many-objective optimization framework for navigating complex fairness and performance trade-offs in machine learning. The approach consists of three main stages: 1) Many-Objective Optimization to explore trade-offs between a candidate set of fairness and performance metrics, 2) Fairness Metric Selection \& Risk Identification to incorporate domain knowledge, human preferences, and insights from Pareto front analysis to prioritize metrics and assess fairness conflict risks, and 3) Many-Objective Model Selection to choose a final model that balances multiple objectives based on assigned weights.}
    \label{fig:flowchart}
\end{figure*}

\begin{abstract}
Fairness-aware Machine Learning (FairML) applications are often characterized by complex social objectives and legal requirements, frequently involving multiple, potentially conflicting notions of fairness. Despite the well-known Impossibility Theorem of Fairness and extensive theoretical research on the statistical and socio-technical trade-offs between fairness metrics, many FairML tools still optimize or constrain for a single fairness objective. However, this one-sided optimization can inadvertently lead to violations of other relevant notions of fairness. In this socio-technical and empirical study, we frame fairness as a many-objective (MaO) problem by treating fairness metrics as \textit{conflicting objectives}. We introduce \textit{ManyFairHPO}, a human-in-the-loop, fairness-aware model selection framework that enables practitioners to effectively navigate complex and nuanced fairness objective landscapes. ManyFairHPO aids in the identification, evaluation, and balancing of fairness metric conflicts and their related social consequences, leading to more informed and socially responsible model-selection decisions. Through a comprehensive empirical evaluation and a case study on the Law School Admissions problem, we demonstrate the effectiveness of ManyFairHPO in balancing multiple fairness objectives, mitigating risks such as self-fulfilling prophecies, and providing interpretable insights to guide stakeholders in making fairness-aware modeling decisions.
\end{abstract}

\section{Introduction}

% Fairness is a key concern
Instances of algorithmic discrimination are a growing concern in both the machine learning (ML) literature and, more recently, in the media and greater society. This is a consequence of the increasing prevalence of ML applications where individuals are disparately impacted by algorithmic decisions \citep{Angwin-propublica16, dezwart-rtc22}. Mirroring the complex and socio-technical nature of the machine bias problem, the field of Fairness-aware Machine Learning (FairML) has emerged, providing a collaborative space for political philosophers, social scientists, legislators, statisticians, and ML researchers. FairML has the overarching goals of defining, studying, detecting, and mitigating algorithmic bias.

% Current FairML approaches oversimplifying and black-box
Despite significant advancements, the FairML community has received widespread criticism from the Social Sciences, Humanities, and Law for attempting to solve the complex, nuanced, and socio-technical problem of machine bias \textit{algorithmically} \citep{Hoffmann-icso19,Selbst-fat19}. Several arguments cite that real-world applications of FairML are often characterized by a complex set of social objectives and legal requirements \citep{Ruf-arxiv21}. Due to their complexity, these criteria are unlikely to be captured by single, coarsely-grained statistical measures of fairness. In such cases, FairML methods that only incorporate a single notion of fairness risk \textit{fair-washing}, or proposing a so-called \textit{fair} model that satisfies one notion of fairness while violates another potentially relevant one, potentially resulting in negative social consequences. Other criticisms of FairML cite the black-box nature of bias-mitigation techniques as a key concern \citep{robertson-aies22}, suggesting a crucial interplay between fairness, transparency, and interpretability \citep{Barocas-nips17, schoffer-kit23}. Broadly effective FairML methods should not only cope with a diverse set of objectives and requirements but ideally offer interpretable insights.

% MOO offers flexibility and transparency
Rather than resisting these criticisms, we embrace them, taking the perspective that FairML approaches that oversimplify the complex and socio-technical nature of FairML problems risk doing more harm than good. Instead, FairML approaches must adapt to the context in which they exist. In recent years, the topic of fairness has gained popularity in the Automated Machine Learning (AutoML) literature, which typically formulates fairness as a bi-objective (BiO) or constrained hyperparameter optimization (HPO) problem. Fairness-aware HPO varies common ML design decisions (tree depth, neural architecture, neural network width, etc.) to explore the Pareto Front of fair and accurate models \citep{wu-corr21,peronne-aies21,schmucker-nips20,dooley-neurips23a}. According to \citet{weerts-arxiv23a}, fairness-aware AutoML holds key advantages in human-centricity and transparency, enabling practitioners to explore multiple fairness-accuracy trade-offs and gain interpretable insights into the fairness-objective landscape of the problem at hand. In addition, the BiO problem formulation extends naturally to the MaO case, encompassing three or more fairness and accuracy constraints or objectives, enabling practitioners to explore not only fairness-accuracy trade-offs but trade-offs between fairness metrics themselves. Although the prospect of constraining or optimizing for multiple notions of fairness has been mentioned several times in the literature \citep{peronne-aies21, schmucker-nips20}, previous studies lack a convincing socio-technical and empirical basis to answer the crucial question: ``\textit{how} and \textit{why} should we put this into practice?" To bridge these gaps in the literature, our study motivates the MaO problem formulation for fairness, making the following contributions:

\begin{enumerate}

    % ManyFairHPO
    \item We propose a human-in-the-loop optimization framework for fairness, \emph{ManyFairHPO}, that aids practitioners in the identification, evaluation, and balancing of fairness metric conflicts and their related social consequences, enabling fairness-aware model-selection decisions that encapsulate multiple conflicting fairness objectives.
    % Evaluation
    \item In an empirical and socio-technical study, we evaluate and motivate the MaO HPO problem formulation for fairness on a diverse set of fairness benchmarks. The key message of our evaluation is ManyFairHPO performs competitively to the typical BiO problem formulation and state-of-the-art bias-mitigation techniques. Additionally, we find that ManyFairHPO thoroughly explores the trade-offs between fairness metrics themselves.
    % Case study
    \item Exemplifying the efficacy of ManyFairHPO in fairness problems with a complex set of fairness objectives, we perform a case study on the topical Law School Admissions problem. We provide a simulation with multiple stakeholders, demonstrating how our interpretable metrics and visualizations can help them understand the potential social consequences of fairness metric conflicts, such as the risk of self-fulfilling prophecy. We show how the interpretability of ManyFairHPO can lead to a thorough, stakeholder-informed model selection decision that mitigates this risk.
\end{enumerate}

In the remainder of the paper, after summarizing related works that incorporate multiple fairness objectives and constraints (Section \ref{section:related-work}), we introduce the related terminology in algorithmic fairness as well as multi-objective optimization (Section \ref{section:background}). We then introduce our approach, ManyFairHPO, and our experimental setup (Section \ref{section:methodology}). Our results (Section \ref{section:results}) show that ManyFairHPO is both suitable as a bias-mitigation technique and enables practitioners to navigate and explore complex fairness objective landscapes. Finally, we conclude with possible future directions of research and a call to action for the fairness community to consider fairness metric conflicts and the MaO problem formulation as a crucial next step in fairness research.

\section{Related Work}
\label{section:related-work}

% Compare to CO
The characterization of fairness as an inherently BiO problem has received significant attention in the literature \cite{martinez-pmlr20, islam-aies21}, with several works citing the importance of considering fairness-accuracy trade-offs in the FairML problem landscape. In this section, we discuss the critical issue of optimization problem formulation for fairness, comparing the typical BiO framing to its counterpart, constrained optimization (CO).

% BO, MILP, and EGR
We begin by discussing several works that formulate the FairML task as a CO problem and discuss its pros and cons compared to the BiO problem formulation. In Fair Bayesian Optimization, \citet{peronne-aies21} expand upon common bias-mitigation techniques to integrate constraints across multiple fairness metrics. Relatedly, \citet{hsu-nips22} employ Mixed Integer Linear Programming to enforce constraints across various fairness criteria, to explore and push the boundaries. Finally, Exponentiated Gradient Reduction (EGR) claims to ``yield a randomized classifier with the lowest (empirical) error subject to the desired constraints" \citep{agarwal-pmlr18}. Due to its strong reputation as a state-of-the-art constrained approach, we select EGR as a baseline in our empirical analysis (Section \ref{section:results}).

% Assumption of prior knowledge
The first key difference between the BiO and constrained problem formulation is that the latter assumes the knowledge of appropriate fairness metrics and achievable constraints. Although previous socio-technical work \citep{Ruf-arxiv21} provides useful guidelines for fairness metric selection, the range of achievable fairness metric values (especially in the presence of conflicts) is highly data and model-dependent. While we align with the argument of \citet{peronne-aies21} that COtimization is more computationally efficient in cases where fairness objectives and reasonable constraints are clearly defined, we maintain that this is an unrealistic scenario, especially when constraining across multiple, potentially conflicting notions of fairness. We also highlight that preferences towards certain fairness metrics or the enforcement of fairness constraints can still be performed after BiO optimization, making it a more flexible method when computational resources are not extremely limited and prior knowledge of the fairness objective landscape is scarce.

% Performance and transparency
Secondly, because constrained approaches focus the search on the constrained region of interest, they typically do not explore the entire Pareto Front of all objectives, often leaving quality (in this case either in terms of performance or fairness) on the table \cite{weerts-arxiv23a}. Additionally, by exploring the entire Pareto Front(s),  approaches provide interpretable insights into the overall objective landscape, a key aspect in building trust \citep{schoffer-kit23} and aiding practitioners in the iterative FairML design cycle \citep{weerts-arxiv23a}. Such insights include but are not limited to the degree of trade-offs between fairness metrics, the nature and shape of Pareto Fronts (e,g. knee points) \citep{dhiman-knrvea19}, and the relative difficulty of different fairness objectives \citep{jordan-nips18}.

% Extensibility
A final key feature of the BiO problem formulation is its natural extension to the MaO setting, which enables practitioners to incorporate multiple user-defined fairness metrics as additional objectives. Although \citet{schmucker-nips20} argue that their methodology is ``extensible" to the MaO setting, they do not verify this hypothesis or explore its benefits. In this study, we fill this gap in the FairML literature, providing a socio-technical and empirical basis for the MaO problem formulation.

\section{Background}
\label{section:background}

\subsection{Fairness-aware Machine Learning}
\label{section:fairness}

In this section, we provide an introduction to fairness and its many definitions, bias mitigation strategies, as well as the theoretical trade-offs between fairness and performance objectives. For a more comprehensive review of the fairness literature, we refer to \citep{pessach-csur22}.

FairML approaches seek to detect and mitigate machine bias at various stages of the ML pipeline and can be effectively divided into pre-processing, in-processing, and post-processing techniques, which detect and mitigate bias in either 1) the input data, 2) the training algorithm or 3) the model’s predictions \citep{pessach-csur22}. In recent years, the objective of fairness has gained increased popularity in the AutoML community, resulting in several studies that achieved competitive performance to specialized bias-mitigation techniques by simply varying the hyperparameters or neural architecture of ML models \citep{peronne-aies21,schmucker-nips20,dooley_neurips23a}. For an in-depth discussion on fairness-aware AutoML, we refer to \citet{weerts-arxiv23a}.

Despite the strong performance of bias-mitigation techniques and fairness-aware AutoML, a lingering question in the fairness literature remains an introspective one: how do we define fairness itself? Grappling with a philosophical question that has been debated for millennia \citep{binns-fact18}, as well as a growing set of social objectives and legal requirements, the FairML community has proposed over 20 different fairness metrics \citep{Barocas-nips17}, which seek to quantify the fairness (or unfairness) of a model. The FairML problem is formally defined below.
\begin{definition}[FairML Problem]
    Given a data set $\mathcal{D} = (X, Y, A)$ of $n$ features and $m$ samples $X \in R^{m\times n}$, a binary target $Y \in \{0, 1\}^m$, and a binary protected attribute $A \in \{0, 1\}^m$, find a model $\mathcal{M}: X \rightarrow \hat{Y}$ that satisfies or optimizes a set of fairness and performance objectives or constraints $\{f_0, f_1, f_2, ...\}$
\end{definition}
The three most common metrics for group fairness are Demographic Statistical Parity (DDSP), Equalized Odds (DEOD), and Equal Opportunity (DEOP). Group fairness operates upon the \textit{egalitarian} principle that valuable resources should be distributed equally across salient demographic groups, while individual fairness is based on the notion of \textit{just deserts} \citep{binns-fact18}, requiring that individuals receive their deserved outcome. Associated individual fairness metrics are similarity-based and require that similar individuals (based on a set of \textit{legitimate factors}) receive similar outcomes \citep{Barocas-nips17}. Due to the challenge in defining legitimate factors, we opt for a simplified definition of Inverse Distance (INVD) as proposed by \citet{berk-corr17}, which treats the ground truth as a legitimate factor (despite the potential for direct bias). We provide a mathematical definition of the above fairness metrics in (Table \ref{table:metrics}).

A foundational concept, especially in the fairness-aware AutoML literature is the \textit{fairness-accuracy trade-off}, which states that improvements in fairness usually come at the cost of reduced predictive accuracy \citep{pessach-csur22}. This is because unfairness is caused primarily by \textit{indirect bias}, which occurs due to an indirect influence of protected attributes on the target variable via a set of \textit{indirect encodings} \citep{kamashima-mlkd12} which may also influence the outcome of interest. For example, standardized test scores are a partially useful predictor of student success. However, these scores can also be indirectly influenced by race via latent factors such as access to test-preparation resources. For a more mathematical discussion of the fairness-accuracy trade-off, we refer to \citep{cooper-aies21}.

\begin{table*}[t]
    \begin{minipage}{\linewidth}
      \centering
        \begin{tabular}{lc}
        \toprule
        Name & Definition  \\
        \midrule
         Demographic Statistical Parity (DDSP)  & $|P(\hat{Y} = 1 | A = 0) - P(\hat{Y} = 1 | A = 1)|$  \\
        Equalized Opportunity (DEOP) &  $|P(\hat{Y} = 1 | A=0, Y=1) - P(\hat{Y} = 1 | A=1, Y=1)|$ \\
         %Equalized (Average) Odds (DEOD) & $\sum_{y\in\{0, 1\}} \frac{ | P(\hat{Y} = 1 | A=0, Y=y) - P(\hat{Y} = 1 | A=1, Y=y)}{2} |$  \\
         Equalized (Average) Odds (DEOD) & $ \frac 1 2 \sum_{y\in\{0, 1\}} \cdot  | P(\hat{Y} = 1 | A=0, Y=y) - P(\hat{Y} = 1 | A=1, Y=y) |$  \\
         %Inverse Distance (INVD) & $\sum_{i, j} \frac{|y_i - y_j| |\hat{y}_i - \hat{y}_j|}{m^2}$ \\
         Inverse Distance (INVD) & $\frac 1 {m^2} \cdot\sum_{i, j=1}^m  |y_i - y_j| \cdot  |\hat{y}_i - \hat{y}_j|$ \\
         \midrule
      \end{tabular}
    \end{minipage}
    \\
    \caption{\textit{Fairness metrics}: Given protected attributes $A$, targets $Y$, and predictions $\hat{Y}$, we summarize measures of unfairness drawn from the \texttt{aif360} library. INVD is a simplified version of similarity-based individual fairness which treats ground truth labels as legitimate factors for the inverse distance function ($|y_i - y_j|$).}
    \label{table:metrics}
\end{table*}

\subsection{Fairness Metric Impossibility Theorem}

A central concept to this study in particular is the \textit{Impossibility Theorem of Fairness}, which states that the three main notions of group fairness (Sufficiency, Separation, and Independence) cannot all be satisfied at once \citep{miconi-arxiv17}, \citep{kleinberg2016inherenttradeoffsfairdetermination} \citep{chouldechova2016fairpredictiondisparateimpact}. The Impossibility Theorem applies when there is a difference in base rates, and is thus widely applicable in real-world FairML problems. According to \citet{Barocas-nips17}:

\begin{quote}
    ``What this shows is that we cannot impose multiple criteria as hard constraints. This leaves open the possibility that meaningful trade-offs between these different criteria exist."
\end{quote}

\subsection{Fairness Metric Conflict-Related Risks}

\begin{figure}[h!]
    \centering
    \includegraphics[width=\linewidth]{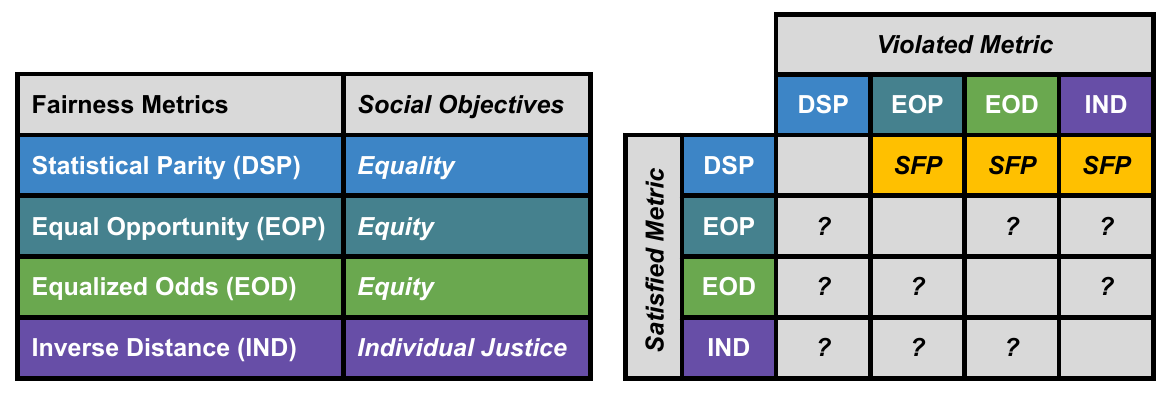}
    \caption{\textit{Conflict-Related Risks}: Fairness metrics serve as a statistical means to specify and optimize various social objectives, such as equality, equity, and individual justice. However, certain conflicts hold potential downstream consequences, such as the risk of Self-Fulfilling Prophecy (SFP) when DDSP is satisfied by violating other metrics.}
    \label{fig:risks}
\end{figure}

Given the potential conflicts between fairness metrics themselves, we formalize the notion of conflict-related risks. In Figure \ref{fig:risks} (left) we first outline our perspective on fairness metrics as a mapping to specific social objectives such as equity, equality, and individual justice. When one metric is satisfied by violating another (right), a \textit{first-order} risk occurs, namely the reduction in the social benefits of the violated metric (e.g. reduced individual justice). However, more consequential are the \textit{second-order} risks of fairness metric violations, which occur rather as a downstream effect. In their critical analysis of DDSP, \citep{dwork-corr11} describe three conflict-related risks, namely 1) reduced utility 2) self-fulfilling prophecy (SFP) and 3) subset targeting. In the context of this study, we focus on the risk of SFP due to its clear connection to a conflict between group fairness metrics DDSP and DEOP. We explain SFP with an example.

Imagine a university admissions algorithm that obtains equal acceptance rates across demographic groups by elevating the chances of underprivileged students getting accepted. This approach leads to the inclusion of several underqualified students from this group. Although this algorithm achieves low DDSP unfairness by achieving equal acceptance rates, it violates the unfairness metric DEOP, which measures the between-group acceptance rate of qualified applicants. In addition to the reduction in equality of opportunity, the second-order risks are more dangerous. If the underprivileged students who are accepted do not succeed at the university, such an admissions strategy could create positive feedback loops for future discrimination, ultimately reinforcing the self-fulfilling prophecy that the underprivileged group is less qualified \citep{dwork-corr11}.

We emphasize, through the various question marks in Figure \ref{fig:risks}, that fairness metric conflict-related risks is a largely understudied topic. Additionally, we note that these conflicts (and their associated risks) are not symmetric, a phenomenon that we describe in Appendix Section A.2.

\subsection{Multi-Objective Optimization}

Multi-objective optimization is an increasingly relevant topic in ML research, as ML applications are increasingly scrutinized with respect to environmental, logistical, and social requirements \citep{karl-arxiv22a}. Multi-objective optimization seeks to provide solutions to an optimization problem under a set of two or more \textit{conflicting} objectives \citep{sharma-acme22}. Objectives are called conflicting when gains in one objective trade-off against losses in the other.

Let $\Lambda$ be a design space representing all possible design decisions. The goal of multi-objective optimization is to find a set of design solutions $\lambda \in \Lambda$ that minimize a multi-criteria objective function $f : \Lambda \rightarrow R^d$, which returns a vector of costs $\langle f_1(\lambda), f_2(\lambda), ..., f_d(\lambda)\rangle$ with respect to each objective. We consider the strict partial Pareto order $\prec_\text{Pareto}$ where $x  \prec_\text{Pareto} y$ iff $x_i \le y_i$ for all $1 \le i \le d$ and $x_j < y_j$ for at least one $j$. If we consider all possible values $\mathcal{Y}=f(\Lambda)$ we obtain a set of minimal values called the \emph{Pareto Front}.
\begin{align} \label{Pareto front}
	\mathcal{P}(\mathcal{Y}) := \big\{ y \in \mathcal{Y}: \{ y' \in \mathcal{Y}, y'\prec_\text{Pareto}  y\} = \emptyset \big\}
\end{align}

Any value in $\mathcal{P}(\mathcal{Y})$ is called \textit{non-dominated}, i.e., there is no other solution that is as least as good as this value for all objectives but strictly better in at least one of the objectives. The \textit{hypervolume} is a quality indicator and represents the volume of the dominated region of the objective space with respect to a reference point $r$.
\begin{equation}
    \mathcal{H}(\mathcal{P}) := \text{Vol}(\mathcal{P}(\mathcal{Y}))
\end{equation}
In this study, we focus on \emph{normalized hypervolume}, an extension of the concept of hypervolume which scales the empirical value to a (0, 1) range, making it comparable across different optimization tasks. For a further discussion on normalized hypervolume and other multi-objective quality indicators, we refer to \citep{hansen-ieee22}.

MaO optimization is a subcategory of multi-objective optimization and is applicable to high-dimensional problems with three or more objectives. MaO algorithms are conceptually similar to low-dimensional multi-objective algorithms but often implement additional quality-diversity measures to ensure sufficient exploration of all objectives \citep{deb-ieee18}. For simplicity and to avoid confusion, we divide multi-objective optimization into two problem settings: BiO (optimizing for two objectives) and MaO (optimizing for more than two objectives). 

\subsection{Hyperparameter Optimization}

Hyperparameter optimization (HPO) seeks to automate the trial-and-error process of designing and deploying ML models \citep{bischl-dmkd23a}. HPO has been shown to consistently improve performance across a wide variety of ML tasks, with high-profile success stories ~\cite{chen-arxiv18a} and has since become a crucial aspect of the ML design cycle. Another advantage of HPO is its extensibility to the BiO case, where varying ML hyperparameters can have an impact on not only performance, but other ML objectives, such as interpretability, energy efficiency, and namely, fairness \citep{karl-arxiv22a}. In the context of fairness-aware AutoML, recent studies have shown a significant impact of regularizing hyperparameters on commonly used fairness metrics. Fairness-aware HPO provides a convenient framework for the FairML problem and is also extensible to include preprocessing, in-processing, and post-processing techniques (as well as their own hyperparameters) as part of the search \citep{wu-corr21}.

\subsection{Evolutionary Algorithms}

Many design spaces are large and non-differentiable, rendering an exhaustive search or gradient-based methods computationally infeasable. A popular approach to multi-objective problems is the Evolutionary Algorithm (EA), a population-based optimization technique that draws inspiration from the process of biological evolution to solve black-box, non-differentiable optimization problems \citep{eiben-springer15}. By implementing bio-evolutionary concepts such as selection, mutation, and crossover, EAs effectively balance exploration and exploitation, generating state-of-the-art results in a variety of domains.

Nondomoninated-Sorting Genetic Algorithm (NSGA-II) is a state-of-the-art MO-EA \citep{deb-ieee02}, which applies the notion of dominance in order to recursively divide the population into ranked fronts. Parent and survivor selection is performed using the heuristics of front rank and crowding distance, which measures the distance of solutions to their nearest neighbor on their respective front. Entire fronts are greedily selected until including the next front exceeds a predefined quota. At this stage, individuals are selected from the final front based on crowding distance, which serves as a heuristic for uniqueness, and encourages exploration through population diversity. NSGA-III \citep{deb-ieee18} incorporates the concept of reference directions in order to select individuals for crossover and survival from equally spaced regions of the many-dimensional objective space.

\section{Methodology}
\label{section:methodology}
In this section, we introduce ManyFairHPO, our MaO optimization approach for fairness that provides an interpretable, human-in-the-loop framework to help practitioners navigate the complex landscape of fairness metric conflicts and make informed and socially responsible model selection decisions. ManyFairHPO consists of three main stages: 1) MaO optimization for a candidate set of fairness metrics, 2) fairness metric selection and risk identification, and 3) fairness-aware model selection. The description below follows Figure \ref{fig:flowchart}, which provides a visual overview of the ManyFairHPO framework.

\subsection{Many-Objective Optimization}
ManyFairHPO takes a FairML dataset and a non-weighted candidate set of performance and fairness metrics ${f_0, f_1, f_2, ...}$ as input. We then employ the popular MaO optimization algorithm NSGA-III \cite{deb-ieee18} to efficiently explore the objective space of fairness-accuracy trade-offs and fairness metric conflicts, yielding a MaO Pareto Front $P(\mathcal{Y}_{MULTI})$. Starting with MaO optimization, ManyFairHPO differs from other approaches by delaying the selection and prioritization of performance and fairness objectives until the trade-offs between these metrics are known.

\subsection{Fairness Metric Selection \& Risk Identification}
The second stage of ManyFairHPO focuses on assigning a set of performance and objective weights $\langle w_0, w_1, w_2, ...\rangle$ based on domain knowledge, human preferences, and insights taken from the high-dimensional Pareto analysis.

\subsubsection{Domain Knowledge and Human Preferences}
Fairness objective weights can be derived from discussions about the relative importance of fairness objectives. To facilitate these discussions, we draw an analogy between fairness metrics and their related social objectives (Section \ref{section:fairness}), making it easier for domain experts and stakeholders to express their preferences. It is important to note that stakeholders and domain experts may have different priorities, which can be accounted for during objective weighting. For instance, a hiring agency might prioritize individual fairness to maintain a reputation for non-discriminatory decisions, while a hiring discrimination expert might focus on group fairness to ensure workplace diversity.

Another crucial discussion point is the identification of potential fairness metric conflict-related risks, which are the potential downstream consequences when one fairness metric is satisfied by violating another (e.g., self-fulfilling prophecy). Key questions to consider include: ``Will ML-assisted decisions be used to train future algorithms?" and ``Are individuals from underprivileged groups being thoughtfully selected?" It is important to keep in mind that the identified risks may not be exhaustive or guaranteed to occur, motivating the need for exploratory analysis of the high-dimensional Pareto Fronts.

\subsubsection{Interpretability via Pareto Analysis}
MaO optimization provides interpretability through post-hoc analysis of the objective landscape. ManyFairHPO offers interpretable metrics and visualizations to identify fairness metric conflicts, visualize them, and evaluate their social consequences. To quantify the strength of fairness metric conflicts, we introduce the notion of fairness metric \textit{contrast}:

\begin{definition}[Contrast]
The \textit{contrast} of fairness metric $f_i$ with respect to fairness metric $f_j$ is defined as the difference in normalized hypervolume when optimizing for $f_j$ and $f_i$, respectively:
\begin{equation}
C(f_i, f_j) := \mathcal{H}_{f_j}\big(\mathcal{P}(\mathcal{Y}j)\big) - \mathcal{H}{f_j}\big(\mathcal{P}(\mathcal{Y}_i)\big)
\label{eq:contrast}
\end{equation}
\label{def:contrast}
\end{definition}

It's worth noting that the contrast metric is not symmetric (Appendix Section A.2). A large positive value of $C(f_i, f_j)$ indicates a severe conflict, where optimizing for $f_i$ fails to optimize for $f_j$. A value close to zero suggests a weak conflict, and a negative value points to a \textit{negative} conflict, where optimizing for a different fairness metric is better than optimizing for one directly.

To help practitioners understand fairness metric trade-offs, we provide visualizations related to contrast, such as 3-dimensional ternary plots, which reveal whether conflicts occur at specific accuracy levels. We also recommend comparing the predictive behavior of models selected from similar accuracy levels on two conflicting Pareto Fronts to assess whether the conflict results in a downstream risk.

\subsection{Fairness-Aware Model Selection}
The final stage of ManyFairHPO involves selecting a model that effectively captures social objectives and mitigates conflict-related risks. Insights from the previous stage can be used to assign performance and fairness objective weights. For a pair of conflicting fairness metrics $(f_i, f_j)$, where satisfying $f_i$ violates $f_j$ and leads to unwanted downstream consequences, a higher weight can be assigned to $f_j$ to avoid this risk (Figure \ref{fig:flowchart}).

Given a high-dimensional Pareto Front $P(\mathcal{Y}_{MULTI})$ and a vector of weights $\langle w_0, w_1, ...\rangle$, the MaO model selection task is simplified to:
\begin{equation}
\lambda^* = argmin_{\lambda} \sum w_i \cdot f_i(\lambda)
\label{eq:selection}
\end{equation}
For simplicity, we define Equation \ref{eq:selection} as a linear combination, but more complex functions can be used to incorporate the utility and marginal gains of metrics.

\subsection{Experimental Setup}
Our experimental scope comprises of hyperparameter search spaces from HPOBench \cite{eggensperger-neuripsdbt21a} for Random Forest (RF), XGBoost (XGB), and Multi-Layer Perceptron (NN) models \cite{scikit-learn} (Appendix Table 2) applied to fairness datasets: Bank Marketing, German Credit, Adult Census Income, COMPAS Criminal Recidivism, and Law School Admissions \cite{vanschoren-sigkdd13a} (Appendix Table 3). We use commonly employed fairness metrics from IBM's \texttt{aif360} library \citep{bellamy-18}: Demographic Statistical Parity (DDSP), Equalized Odds (DEOD), Equality of Opportunity (DEOP), and Inverse Distance (INVD) (Appendix Table \ref{table:metrics}).

We apply NSGA-III to optimize for a five-dimensional Pareto Front $P(\mathcal{Y}_{MULTI})$ of $F_1$-Score and all four fairness metrics, resulting in 15 MaO experiments repeated with 10 random seeds (150 total runs). For comparison, we apply BiO optimization (NSGA-II) to optimize for $F_1$-Score and a single fairness metric, resulting in 60 BiO experiments repeated with 10 random seeds (600 total runs). We also compare our results with Exponentiated Gradient Reduction (EGR) by in-processing the most accurate model for each search space and dataset across 10 independent trials. The source code for experiments and analysis as well as supplementary material is available at \url{https://github.com/jr2021/ManyFairHPO-AIES}.

\subsubsection{Research Questions}
To systematically evaluate the effectiveness of ManyFairHPO and demonstrate its practical application, we propose the following research questions:

\begin{itemize}
\item [\textbf{RQ1}:] \textit{Fairness-Aware Many-Objective Optimization.} What are the advantages of ManyFairHPO as an approach to bias mitigation compared to BiO optimization and state-of-the-art bias-mitigation techniques? How well does it explore fairness-accuracy trade-offs and fairness metric conflicts? What interpretable insights can be gained into fairness objective landscapes that can be used to select and prioritize fairness metrics, as well as identify and assess conflict-related risks?
\item [\textbf{RQ2}:] \textit{Law School Admissions Case Study.} How can ManyFairHPO be applied in a real-world scenario with complex fairness objectives and diverse stakeholder interests? We explore a Law School Admissions case study to demonstrate how ManyFairHPO can relate social objectives and downstream risks to a prioritized set of fairness metrics, informing the selection of a model that incorporates multiple conflicting performance and fairness objectives.
\end{itemize}

\section{Results}
\label{section:results}

\subsection{ManyFairHPO for Bias-Mitigation}
In this section, we systematically assess the effectiveness of the MaO problem formulation (ManyFairHPO) from a technical perspective. We demonstrate that ManyFairHPO is competitive from a bias-mitigation standpoint, achieving similar fairness and accuracy values compared to BiO optimization and outperforming the state-of-the-art bias-mitigation technique EGR. Moreover, we showcase ManyFairHPO's novel ability to explore and balance trade-offs between fairness metrics themselves.

\subsubsection{Fairness-Accuracy Trade-offs}
We first verify that optimizing for multiple fairness metrics together using ManyFairHPO is at least as effective as optimizing for the same metrics separately using BiO optimization.

\begin{figure}[h!]
\centering
\includegraphics[width=\linewidth]{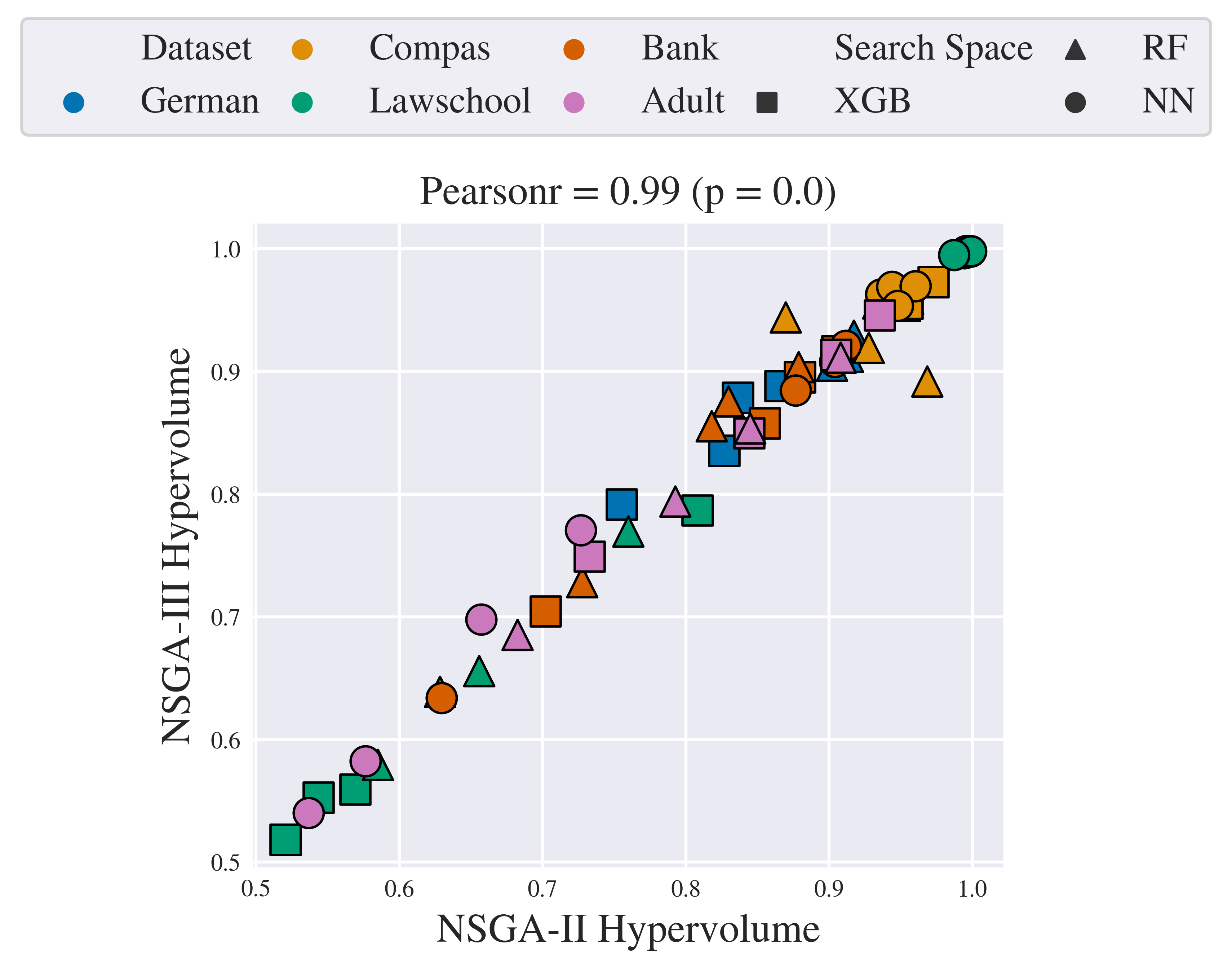}
\caption{\textit{ManyFairHPO vs. Bi-Objective Optimization}: Comparison of hypervolume achieved by ManyFairHPO and bi-objective optimization across datasets and search spaces. NSGA-III matches the hypervolume achieved by NSGA-II with a correlation of 0.991, indicating that similar fairness-accuracy Pareto Fronts are achieved when optimizing for fairness metrics together compared to optimizing for them separately.}
\label{fig:hv_regret}
\end{figure}

\setlength{\tabcolsep}{0.375em}
\begin{table}[h!]
\begin{minipage}{\linewidth}
\centering
\begin{tabular}{llccc}
\cmidrule(r){1-5}
&  & & Search Space & \\
\cmidrule(r){1-5}
Dataset & Objectives & XGB & RF & NN \\
\midrule
German & F1-DDSP & \textbf{10/0/0} & \textbf{10/0/0} & \textbf{10/0/0} \\
& F1-DEOD & \textbf{10/0/0} & \textbf{10/0/0} & \textbf{10/0/0} \\
Compas & F1-DDSP & \textbf{10/0/0} & 0/10/0 & \textbf{10/0/0} \\
& F1-DEOD & 5/5/0 & 0/0/10 & \textbf{10/0/0} \\
Lawschool & F1-DDSP & 0/10/0 & 0/0/10 & \textbf{10/0/0} \\
& F1-DEOD & 0/0/10 & 0/0/10 & \textbf{7/3/0} \\
Bank & F1-DDSP & \textbf{10/0/0} & \textbf{10/0/0} & \textbf{10/0/0} \\
& F1-DEOD & \textbf{10/0/0} & \textbf{10/0/0} & \textbf{10/0/0} \\
\midrule
\end{tabular}
\end{minipage}
\
\caption{\textit{ManyFairHPO vs. In-Processing (EGR)}: Number of wins (ManyFairHPO dominates EGR), ties (non-dominated solutions), and losses (EGR dominates ManyFairHPO) across different datasets and search spaces. ManyFairHPO dominates EGR in 9/12 cases for DDSP and 7/12 cases for DEOD.}
\label{table:egr}
\end{table}

\begin{figure*}[t!]
\centering
\includegraphics[width=\textwidth]{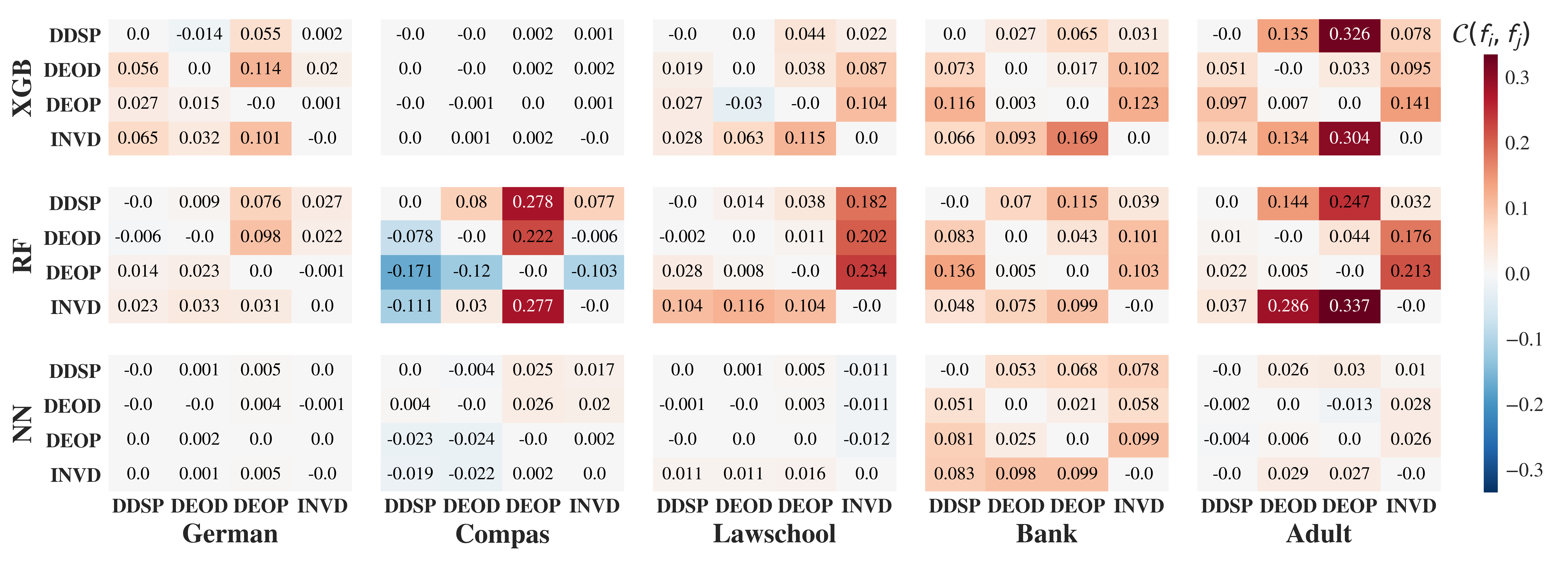}
\caption{\textit{Fairness Metric Contrast}: Heatmap visualization of fairness metric conflicts across different datasets and ML models, measured by the degree to which optimizing for one fairness-accuracy trade-off fails to optimize for another. Light red cells indicate more severe conflicts. The results highlight the data and model dependence of fairness metric conflicts.}
\label{fig:contrast}
\end{figure*}

To empirically validate this, we compare the fairness-accuracy hypervolume (normalized) achieved by the MaO optimizer NSGA-III compared to its BiO counterpart NSGA-II  across different BiO combinations, datasets, and hyperparameter search spaces (Figure \ref{fig:hv_regret}). We observe a strong and significant correlation of 0.991 ($p$ = 0.0) between the hypervolume achieved from optimizing for multiple fairness metrics together as opposed to optimizing for them separately.

Next, we compare the Pareto Fronts obtained from ManyFairHPO to EGR, a state-of-the-art bias-mitigation technique that reduces the FairML task to a series of cost-sensitive classification problems \cite{agarwal-pmlr18}. To provide a fair comparison, we in-process the most accurate hyperparameter configurations with EGR for 10 independent trials across all search spaces and models with respect to group fairness metrics DDSP and DEOD.\footnote{Our results do not include a comparison on the Adult dataset because single evaluations of EGR exceeded our 24-hour time budget.} We summarize the wins, ties, and losses (W/T/L) of ManyFairHPO against EGR in Table \ref{table:egr}, where ManyFairHPO dominates EGR in the majority of trials in 9/12 cases for DDSP and 7/12 cases for DEOD. We also visualize these outcomes in Appendix Figure 2. The dominance of ManyFairHPO demonstrates that MaO optimization is capable of efficiently exploring multiple fairness-accuracy trade-offs, achieving comparable Pareto Fronts to BiO optimization and outperforming the state-of-the-art bias-mitigation technique EGR in most cases.

\subsubsection{Fairness Metric Trade-offs}
Beyond achieving competitive fairness-accuracy trade-offs, a key novelty of ManyFairHPO is its ability to explore trade-offs between fairness metrics themselves. To identify which fairness metrics conflict on which search spaces and datasets, we calculate the contrast between fairness metrics $C(f_i, f_j)$ as described in Definition \ref{def:contrast} and provide a summary of observed conflicts in Figure \ref{fig:contrast}. A dark red cell in row $j$, column $i$, indicates that optimizing for $F_1$-Score and fairness metric $f_i$ fails to optimize for $F_1$-Score and fairness metric $f_j$.

A key insight from this analysis is the strong data dependence of fairness metric conflicts, motivating the need for MaO exploration of fairness metric trade-offs on real-world problems. We observe similar conflict patterns on datasets with semantically similar target variables, such as the Bank and Adult datasets, whose targets (Bank Marketing Subscription and Income) both serve as proxies for financial status. This suggests a relationship between FairML problem characteristics and the presence of certain conflicts, motivating future work on which conflicts and risks to be aware of for different classes of tasks. We also observe a slight model dependence, with lower error models displaying stronger conflicts, indicating that ManyFairHPO should be applied over multiple hyperparameter search spaces to reveal the true fairness objective landscape. Finally, we observe several negative conflicts on the RF-Compas experiment, where a fairness metric is better optimized indirectly than directly, a result which suggests a possible interaction effect between fairness metrics in hyperparmeter search spaces. We further explain this result in Appendix A.1.

\begin{figure}[h!]
\centering
\includegraphics[width=\linewidth]{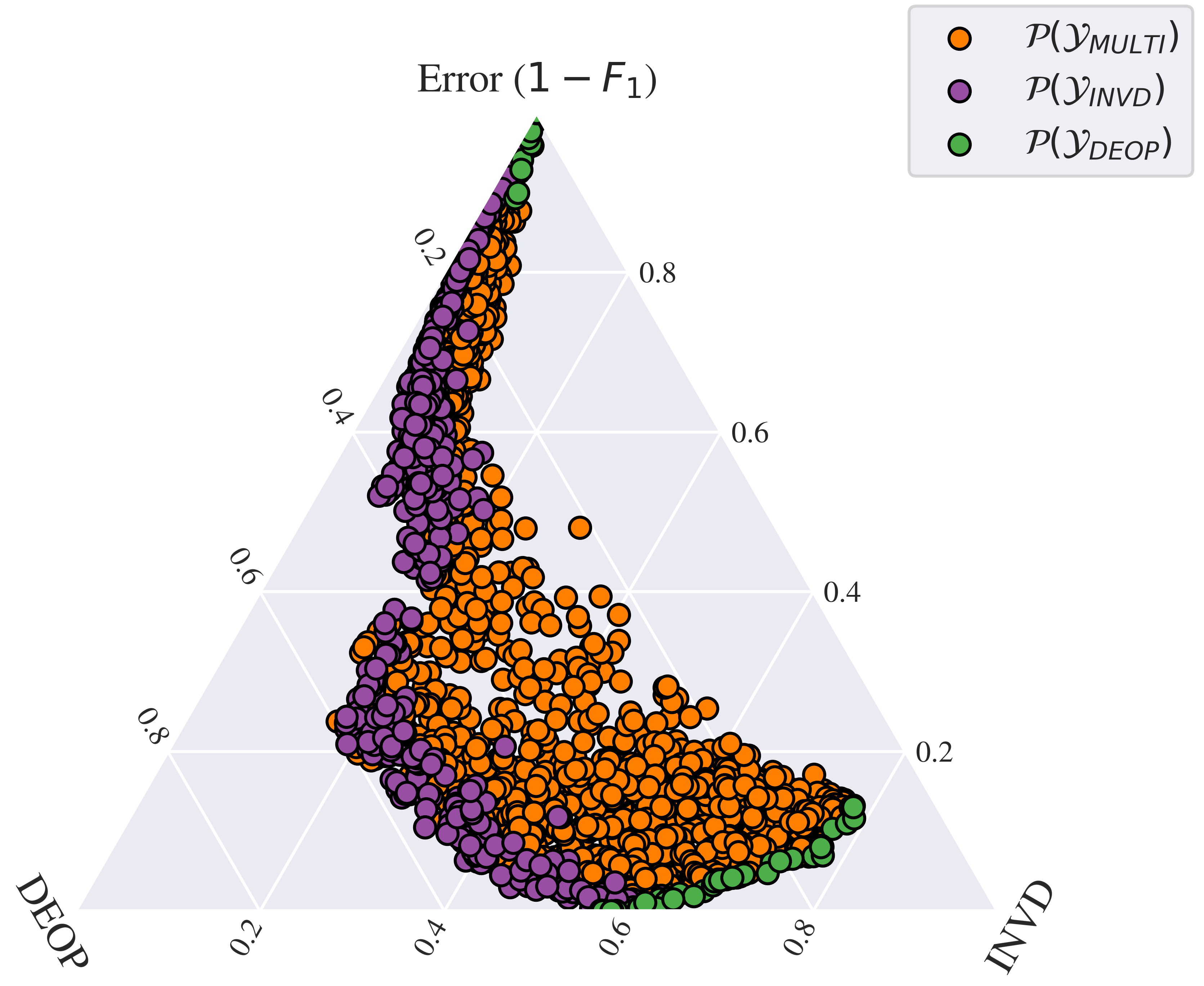}
\caption{\textit{Conflict Interpolation (RF-Adult)}: Ternary plot of the high-dimensional Pareto Front (orange) in the presence of a fairness metric conflict between INVD (purple) and DEOP (green). Metric values get better further away from each corner. The many-objective Pareto Front provides a selection of models that interpolate between fairness metric conflicts.}
\label{fig:many}
\end{figure}

To illustrate ManyFairHPO's novel capability in balancing fairness metric conflicts, we focus on a conflict observed between INVD and DEOP ($C=0.337$) from the RF-Adult experiment (Figure \ref{fig:contrast} top-right). In Figure \ref{fig:many}, we provide a ternary plot of this conflict projecting the 3-dimensional normalized locations of the Pareto Fronts $\mathcal{P}(\mathcal{Y}_{DEOP})$ and $\mathcal{P}(\mathcal{Y}_{INVD})$ and $\mathcal{P}(\mathcal{Y}_{MULTI})$ in 2-dimensional space.

\begin{figure*}[t!]
    \centering
    \includegraphics[width=0.875\linewidth]{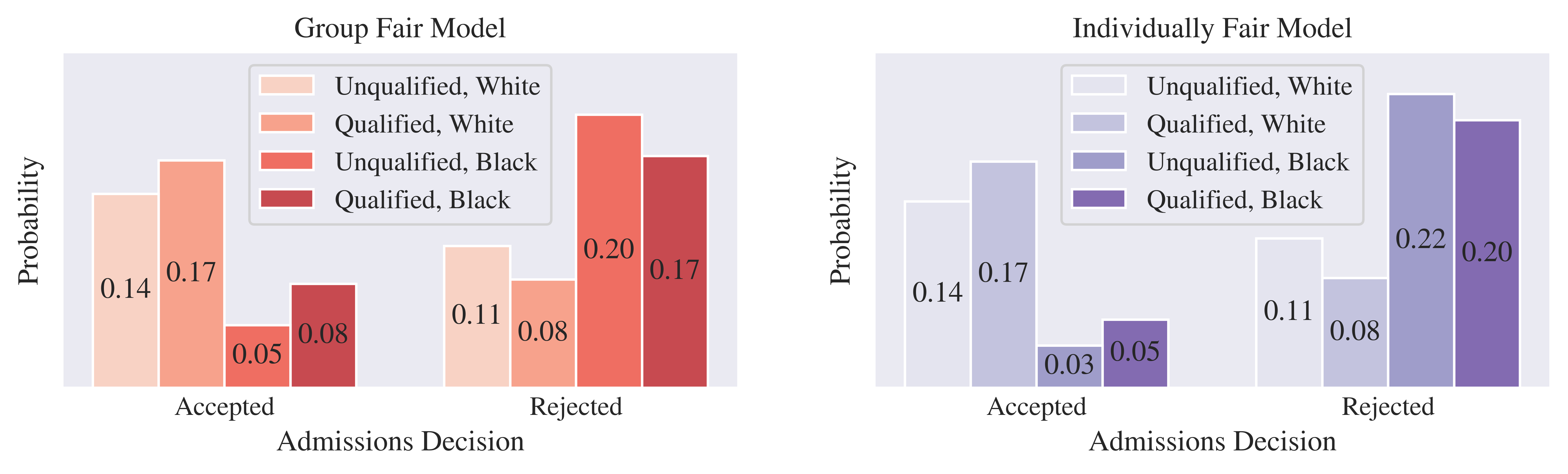}
    \caption{\textit{Risk Assessment (RF-Lawschool)}: Explanation of the conflict between individually-fair and group-fair models discovered selected from the RF-Lawschool experiments. The individually fair model has a higher rejection rate, resulting in fewer similarly unqualified Black and White students receiving different outcomes (individual fairness). However, this strategy also increases between-group acceptance rates (group unfairness).}
    \label{fig:rf-lawschool}
\end{figure*}

\begin{figure}[h!]
    \centering    \includegraphics[width=\linewidth]{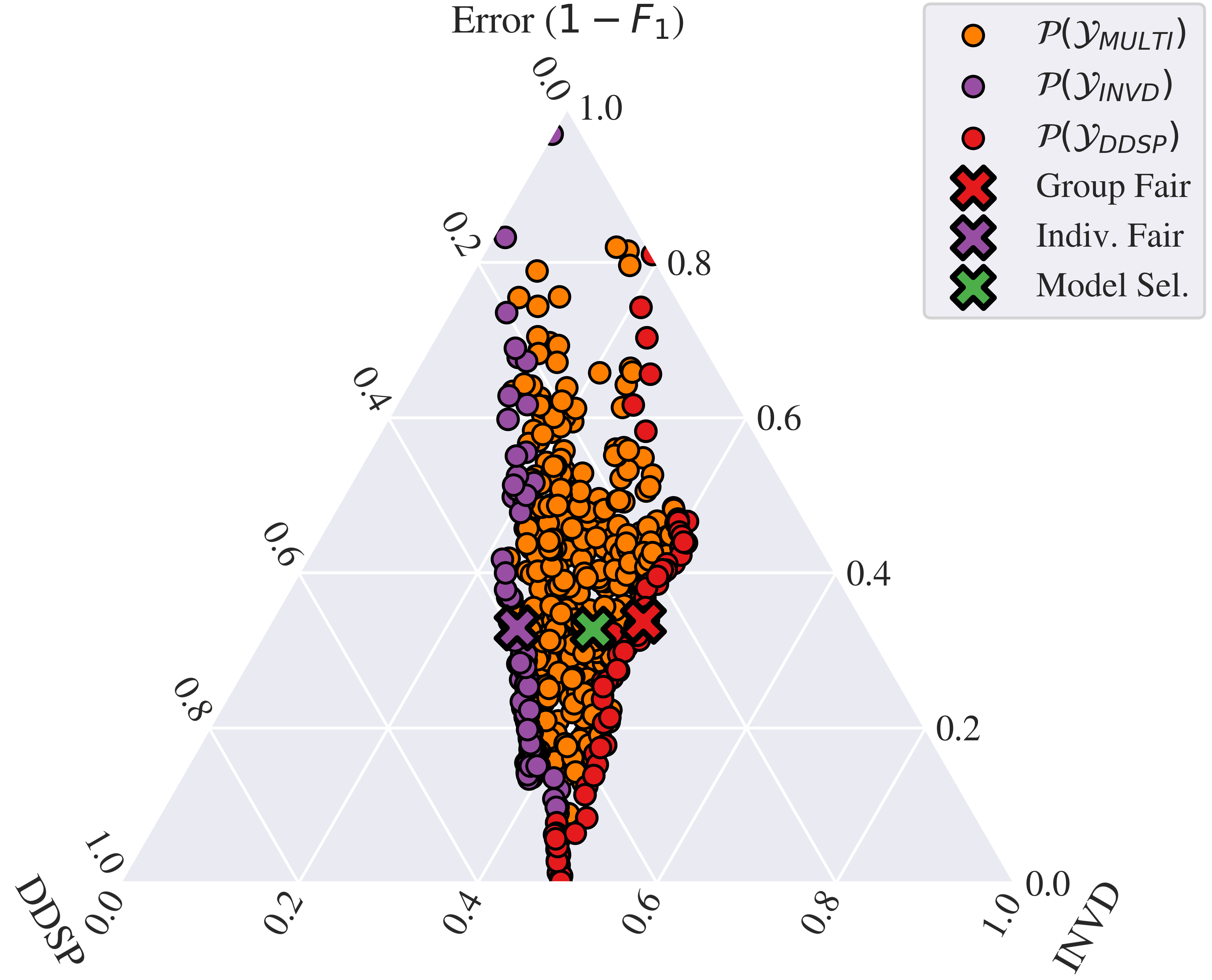}
    \caption{\textit{Model Selection (RF-Law School)}: Model selection on the Law School Admissions problem under objective weights $\mathbf{w}=\langle 0.5, 0.2, 0.3 \rangle$ for $F_1$-Score, DDSP, and INVD. Single objective scalarization leads to a modeling decision (green cross) with strong accuracy ($F_1 = 0.7$) that successfully mitigates the risk of self-fulfilling prophecy.}
    \label{fig:expl_lawschool}
\end{figure}

The first observation we make is that INVD and DEOP are only not in conflict at the high and low error extrema of the objective space, where the model selections are either unusable due to high error (top corner) or perform poorly in terms of INVD and DEOP (bottom edge). In the top 20\% of most accurate models, the Pareto Fronts $\mathcal{P}(\mathcal{Y}_{DEOP})$ and $\mathcal{P}(\mathcal{Y}_{INVD})$ diverge, meaning that for a given accuracy level in this range, a selection from the $\mathcal{P}(\mathcal{Y}_{DEOP})$ Pareto Front would result in a significant trade-off in INVD. In contrast, the MaO Pareto Front $\mathcal{P}(\mathcal{Y}_{MULTI})$ provides a large number of model selections that interpolate between these choices, allowing practitioners to trade-off fairness metrics with a finer grain.

In summary, these results showcase ManyFairHPO's ability to not only find competitive fairness-accuracy trade-offs compared to BiO optimization and state-of-the-art bias-mitigation techniques but also to explore and balance trade-offs between fairness metrics themselves. This is a novel contribution to the FairML literature.

\subsection{Law School Admission Case Study}
To illustrate the practical application of ManyFairHPO in navigating intricate fairness landscapes, we present a case study using the Law School Admissions dataset from the 1991-1997 LSAC National Longitudinal Bar Passage Study \cite{wightman98}.

\subsubsection{Problem Description and Stakeholder Analysis}
This dataset, containing information from nearly 30,000 Law School applicants, is commonly used in the fairness community to simulate acceptance decisions based on predicted first-year average (FYA) scores. Given the high ethnic disparity identified in the initial LSAC study, we demonstrate how ManyFairHPO can be used to balance the complex and conflicting fairness objectives relevant to this scenario.

As outlined in Section \ref{section:methodology}, the first step of ManyFairHPO is MaO optimization. We apply NSGA-III to optimize for performance ($F_1$) as well as fairness metrics DDSP, DEOP/D, and INVD. Before examining the fairness metric conflicts, we introduce a hypothetical scenario involving two stakeholders with different preferences toward higher-level objectives.

The University, the first stakeholder, aims to make admissions decisions that maintain a high academic standard while attracting a large applicant pool. Consequently, they are hesitant towards Affirmative Action strategies that may discourage applicants from well-represented groups. The second stakeholder, the University's Equality, Diversity, and Inclusion (EDI) committee, prioritizes increasing diversity and breaking down systemic inequalities in both the admissions process and the University as a whole. After discussing their objectives, the University and EDI committee agree on initial weights of $\mathbf{w} =\langle 0.5, 0.5 \rangle$ for $F_1$-Score and DDSP, respectively.

\subsubsection{Conflict Evaluation and Assessment}
With the initial objective weights established based on the University and EDI committee's preferences, we demonstrate how ManyFairHPO's interpretability can assist practitioners in identifying, understanding, and assessing fairness metric conflict-related risks.

Figure \ref{fig:contrast} reveals a conflict between the individual fairness metric INVD and group fairness metrics DEOP ($C = 0.234$), DEOD ($C = 0.202$), and DDSP ($C= 0.182$). Notably, the group fairness metrics do not conflict with each other, implying that the assigned weight $w_{DDSP}=0.25$ for DDSP captures the EDI committee's preference for all three group fairness metrics. Additionally, the conflict is slightly asymmetric, meaning that selecting a Pareto optimal model for INVD results in a smaller violation of the EDI committee's preferred group fairness metrics compared to the reverse scenario.

Figure \ref{fig:expl_lawschool} further illustrates this conflict, showing that $\mathcal{P}(\mathcal{Y}_{INVD})$ fails to approximate group fairness Pareto Fronts, particularly in high to moderate error, low group unfairness regions of the $(F_1, DDSP)$ objective spaces (40-50\%).

To exemplify the potential downstream social consequences of this fairness metric conflict, we compare the predictive behavior of two models: one selected from the medium error, moderate individual unfairness region of $\mathcal{P}(\mathcal{Y}_{INVD})$, and another chosen at a similar error level from $\mathcal{P}(\mathcal{Y}_{DDSP})$. The objective space locations of these models are also visualized in Figure \ref{fig:expl_lawschool} as ``Group Fair" and ``Indiv. Fair".

Next, we perform risk-assessment on the two selected models, providing a detailed summary of these models' predictive behavior in Figure \ref{fig:rf-lawschool}. Both models exhibit relatively low group and individual unfairness (but high error) by increasing the predicted qualification rate for Black students from $P(Qualified | Black) = 0.01$ in the data to $P(Accept | Black) = 0.08$ for the individually fair model and $P(Accept | Black) = 0.13$ for the group-fair model. The individually fair model rejects 3\% more qualified and 2\% more unqualified Black applicants compared to the group-fair model. This improves INVD, as 2\% fewer similarly unqualified Black and White applicants receive different outcomes. However, the group-fair model accepts 5\% more Black applicants overall, reducing the disparity between group acceptance rates (DDSP) from 23\% in the individually-fair model to 18\%. While the group-fair model achieves a more diverse accepted class, aligning with the EDI committee's objectives, it does so by accepting 2\% more unqualified Black applicants than the individually-fair model.

Implementing such an admissions strategy may lead to the downstream consequence of \textit{self-fulfilling prophecy} if the underprivileged students admitted through this approach struggle academically, which concerns both the University and the EDI committee.

\subsubsection{Fairness-Aware Model Selection}
Having identified the fairness metric conflict between INVD and group fairness metrics, the University and EDI committee update the weights from $\mathbf{w} = \langle 0.5, 0.5 \rangle$ to include a preference for INVD to mitigate the risk of SFP. The new weights for $F_1$, DDSP, and INVD become $\mathbf{w} = \langle 0.5, 0.2, 0.3 \rangle$. Using these objective weights, we perform model selection via single-objective scalarization (Figure \ref{fig:expl_lawschool}) to perform a model selection (green cross) that achieves a moderate level of accuracy while balancing DDSP and INVD to mitigate the risk of SFP. This result satisfies both the University's and EDI committee's objectives.

This case study shows the effectiveness of ManyFairHPO in capturing and navigating the complex fairness objective landscape of the Law School Admissions problem, considering the preferences and concerns of multiple stakeholders. By providing interpretable insights into fairness metric conflicts and their potential downstream consequences, ManyFairHPO enables the University and EDI committee to make informed decisions that align with their priorities and mitigate risks.

\section{Conclusion}
\label{section:conclusion}

In this study, we address the criticisms of FairML approaches that attempt to solve the fairness problem \textit{algorithmically} by motivating the consideration of fairness metrics as conflicting objectives. Our main contribution is ManyFairHPO, a human-in-the-loop and interpretable MaO optimization framework for fairness that enables practitioners to thoroughly navigate complex fairness objective landscapes. ManyFairHPO is designed to encourage socially responsible fairness modeling decisions that encapsulate multiple fairness objectives and avoid fairness metric conflict-related risks.

Our empirical evaluation demonstrates that ManyFairHPO achieves similar Pareto fronts compared to the typical BiO optimization and even outperforms the state-of-the-art bias-mitigation technique Exponentiated Gradient Reduction (EGR). Moreover, ManyFairHPO exhibits a unique ability to explore and balance trade-offs between fairness metrics themselves, providing interpretable insights into problems-specific fairness objective landscapes. We further illustrate the practical application of ManyFairHPO through a Law School Admissions case study, guiding practitioners from fairness metric selection and risk identification to MaO model selection. The case study showcases ManyFairHPO's effectiveness in navigating complex fairness objective landscapes, considering the preferences and concerns of multiple stakeholders, and mitigating potential downstream consequences.

One key assumption we make is that stakeholder compromise is possible in the given use case, and that a compromising solution can be found by ManyFairHPO. Through our comparison to BiO HPO and EGR we show that ManyFairHPO is largely effective at approximating fairness-accuracy and fairness-fairness Pareto Fronts. However, the ManyFairHPO's approximation largely depends on the quality of the hyperparameter search space. As mentioned in Section \ref{section:background}, future extensions or applications of ManyFairHPO could see benefit from including bias mitigation approaches and their hyperparameters in the search space as seen in \cite{wu-corr21}. From a socio-technical perspective, we note that the success of ManyFairHPO as a socio-technical framework also rises and falls with honest human interaction, including preferences from a diverse set of stakeholders as well as thorough analysis and assessment of fairness metric conflicts and their related social consequences. While we believe ManyFairHPO is a valuable approach, it could have adverse impacts if mitigating one conflict-related risk leads to another unforeseen social consequence. 

% To avoid potential issues, we advocate for in-depth interdisciplinary discussions when using ManyFairHPO in practice.

% Based on our findings, we recommend that both the fairness community and FairML tools begin to recognize and treat fairness metrics as potentially conflicting objectives. FairML techniques should be capable of incorporating and balancing multiple fairness objectives while providing interpretable insights into the fairness-objective landscape.

This work also paves the way for socio-technical discussions on how fairness metrics can be used in balance to meet complex social objectives and requirements while mitigating downstream conflict-related risks, rephrasing the question from ”which fairness metric should I use?” towards ”which prioritization of metrics would balance social objectives and mitigate risks?” We also encourage socio-technical studies to identify further conflict-related risks and form connections between problem characteristics and the presence, severity, and significance of fairness metric conflicts. 

\section*{Acknowledegments}

Funded by the Deutsche Forschungsgemeinschaft (DFG, German Research Foundation)-SFB1597-499552394. The authors of this work would like to thank the reviewers and the the organization of the Artificial Intelligence, Ethics, and Society Conference (AIES '24) for the opportunity to share our work and receive feedback. We also would like to thank Dr. Janek Thomas for his contribution through his supervision of the original Master's Project which this work stemmed from. We would finally like to thank the Zuse School ELIZA Master’s Scholarship Program for their financial and professional support of our main author.

\bibliography{bib/strings,bib/lib,bib/local,bib/proc}

\end{document}

\maketitle

\newtheorem{definition}{Definition}[section]

\appendix

\section{Supplementary Results}
\label{section:supplementary}

\subsection{Criminal Recidivism: Negative Conflicts}
\label{section:compas}

The Compas data set is a seminal instance of algorithmic bias, where a sentencing algorithm used in the Florida judicial system to predict criminal recidivism (likelihood of re-offending) to aid parole decisions was found to be severely biased against Black defendants \citep{Angwin-propublica16}. In this section, we take a closer look into the so-called \textit{inverse conflicts} we observed in the RF-Compas experiment, where optimizing for different fairness metrics found stronger solutions than optimizing for a fairness metric directly. This result suggests a possible explanation for the strong performance of the MaO problem formulation in this scenario (Main Figure 3), where interaction between fairness metrics enables MaO to discover strong overall solutions.

\begin{figure*}[t!]
    \centering
    \includegraphics[width=0.43\linewidth]{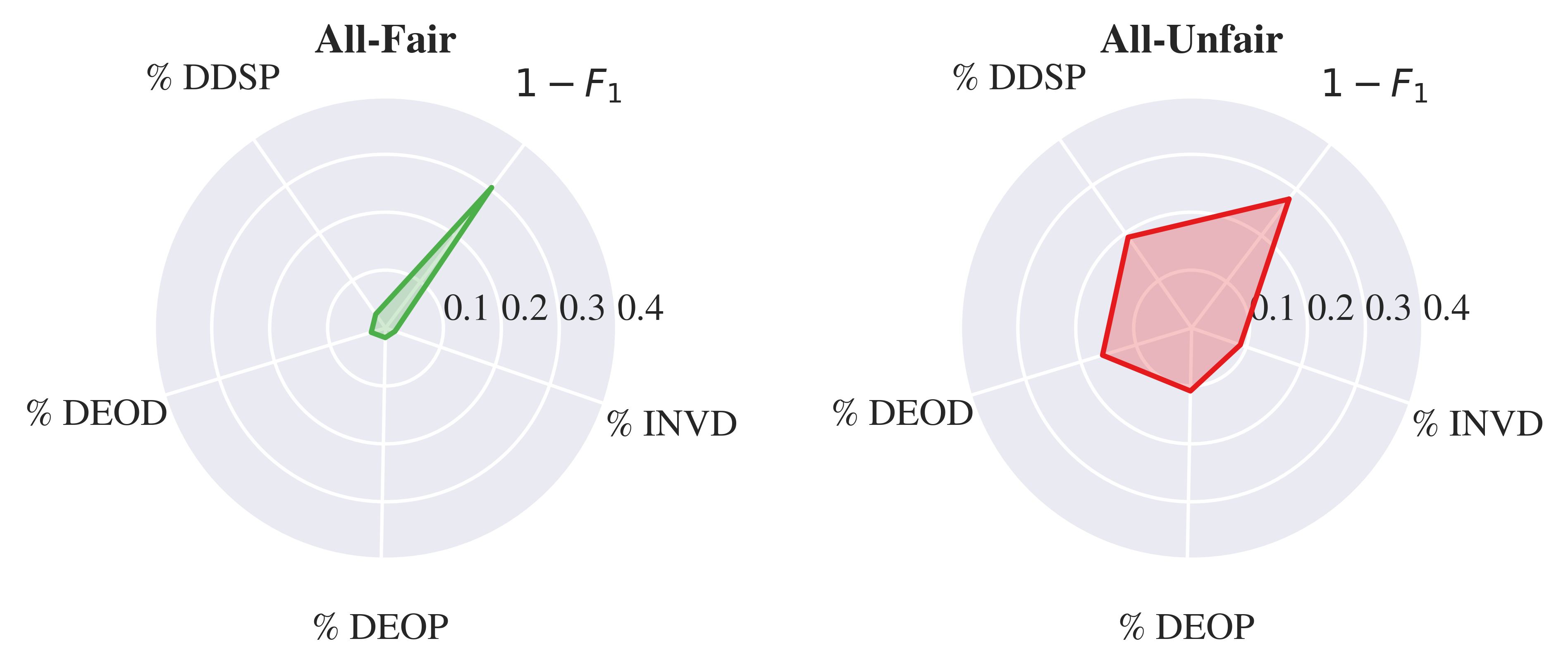}
    \includegraphics[width=0.425\linewidth]{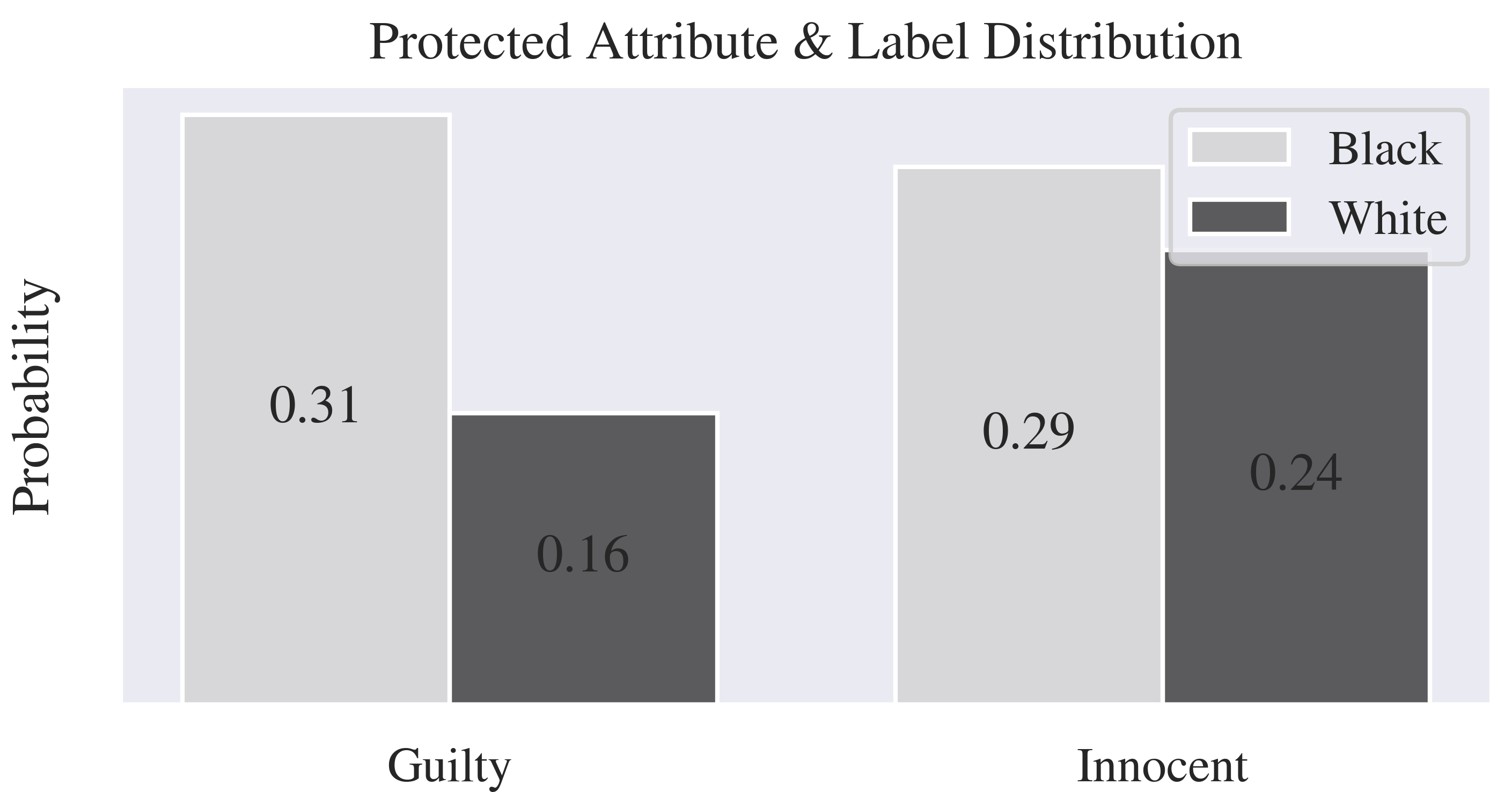}
    \includegraphics[width=0.875\linewidth]{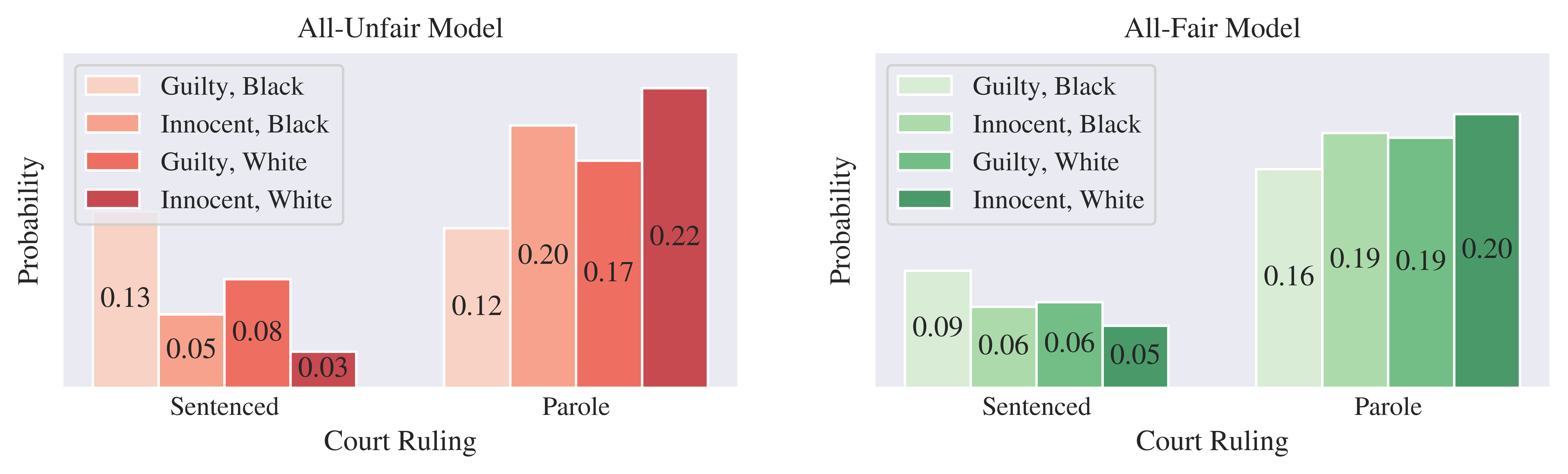}
    \caption{\textit{Inverse Conflicts (RF-Compas)}: Overall-fair model discovered in the RF-Compas experiment, which minimizes all fairness metrics at a reasonable accuracy level. The All-Fair model increases the parole rate for Black defendants that did in fact re-offend to the same rate as White applicants, resulting in low group and individual unfairness.}
    \label{fig:rf-compas}
\end{figure*}

In Main Figure 4 we observe negative contrast values between DDSP with respect to DEOD ($C=-0.078$), DEOP ($C=-0.171$), and INVD $(C=-0.111)$, indicating that a fairer solutions in terms of DEOD/P and INVD were discovered when optimizing for $F_1$-Score and DDSP. We also observe MaO experiments in Main Figure 3 with negative regret (-5\%  to -10\%), indicating that a higher $\mathcal{H}_{DEOD/P}$ and $\mathcal{H}_{INVD}$ was achieved by the MaO experiment than by their corresponding BiO experiments. These results are attributed to a single model discovered on $\mathcal{P}(\mathcal{Y}_{DDSP})$ which achieves reasonable accuracy ($1 - F_1 = 0.3$) and the lowest unfairness in terms of all fairness metrics (Appendix Figure \ref{fig:rf-compas}).

In order to better understand how strong overall fairness was achieved by this model, we compare its behavior with a slightly more accurate ($1-F_1=0.27$) but less fair model in terms of all fairness metrics (Figure \ref{fig:rf-compas}). In comparison, the all-fair model has a lower sentencing rate for Black defendants that re-offended ($P(Sentenced | Black, Guilty) = 0.09$) than the overall-unfair model ($P(Sentenced | Black, Guilty) = 0.13$). However, because both models have a high parole rate for White defendants that re-offend $($$P(Parole | White, Guilty) \geq 0.20$), the decreased sentencing rate from the overall-fair model has the effect of improving overall fairness. First of all, the between-group parole rate $P(Parole | White) - P(Parole | Black)$ is improved from DDSP = 0.07 in the overall-unfair model to DDSP = 0.04 in the overall-fair model. In addition, the between-group parole rate for non-re-offending defendants $P(Parole | White, Innocent) - P(Parole | Black, Innocent)$ is improved from DEOP = 0.02 in the overall-unfair model to DEOP = 0.01 in the overall-fair model. Finally, similarity-based individual fairness INVD is also improved in the overall-fair model, as 4\% more similarly re-offending defendants receive similar parole outcomes.
This result suggests that optimizing for multiple notions of fairness can have the effect of unlocking regions of the objective space that are otherwise inaccessible using the BiO problem formulation.

\subsection{Asymmetry of Fairness Metric Conflicts}
\label{section:asymmetry}

In this section, we outline a scenario where a substantial difference in base rates leads to an asymmetric fairness metric conflict. An asymmetric fairness metric conflict occurs when the impact of satisfying fairness metric $f_i$ on violating fairness metric $f_j$ is different (larger or smaller) from the impact of satisfying $f_j$ on violating $f_i$.

In Main Figure 4, we observe an interesting phenomenon on the RF-Lawschool experiment, where the conflict between INVD with respect to group fairness metrics DDSP and DEOP/D ($C = 0.2$) is significantly larger than the conflict between group fairness metrics DDSP and DEOP/D with respect to INVD ($C = 0.1$). In plainer terms, if INVD is satisfied it leads to a strong violation of DDSP, while the satisfaction of group fairness metrics leads to a relatively weaker violation of INVD.

In the following argument, we explain why asymmetry occurs on the Lawschool dataset by drawing a connection to the significant imbalance (92\% White and 8\% Black) in privileged and unprivileged applicants (Appendix Table \ref{table:dataset}). Consider a perfect classifier that satisfies INVD by accepting all qualified applicants and rejecting all unqualified ones. Referring to the distribution in Main Figure 6 (right), the classifier thus accepts all 1\% of applicants who are qualified and Black as well as all 50\% who are qualified and White. Similarly, the classifier rejects all 7\% of applicants who are unqualified and Black as well as all 42\% who are unqualified and White. Although INVD is satisfied (all individuals receive the outcome they deserve, regardless of demographic group), such an admissions strategy strongly violates DDSP (precisely resulting in $P(Accept | White) - P(Accept | Black) = \frac{42}{92} - \frac{1}{8} = 0.34$ or a difference in between-class acceptance rate of 34\%).

Now consider modifying this classifier (e.g. with a postprocessing technique) such that DDSP is satisfied by increasing the acceptance likelihood for unqualified Black students by 3\% and decreasing the acceptance likelihood for qualified White students by 4\% (positive discrimination), resulting in $P(Accept | White) - P(Accept | Black) = \frac{46}{92} - \frac{4}{8} = 0$. Such a modified classifier results in only a 24\% violation of INVD, as 3\% and 4\% of similarly qualified (or unqualified) applicants from different demographic groups receive different admissions/rejection outcomes.

Note that the impact of DDSP on INVD depends on the base rate of privileged/unprivileged applicants, and asymmetry would increase in this scenario if the \textit{overall} proportion of Black applicants increased while the ratio of qualified and unqualified Black applicants stayed the same. For example, if 2\% of applicants were qualified and Black, while 14\% of applicants were unqualified and Black, satisfying DDSP, would require a 6\% (as opposed to the previous 3\%) increase in acceptance likelihood for unqualified Black applicants, leading to a larger increase in INVD than in the previous example. We thus exemplify how fairness metric conflicts can be asymmetric, while also identifying the impact that dataset characteristics (e.g. difference in base rates) can have on their occurrence, significance, and symmetry. This identification suggests that fairness metric conflicts can potentially be anticipated during domain-knowledge-driven deliberations, adding a technical and concrete angle to these discussions.

\begin{figure*}
    \centering
    \includegraphics[width=\textwidth]{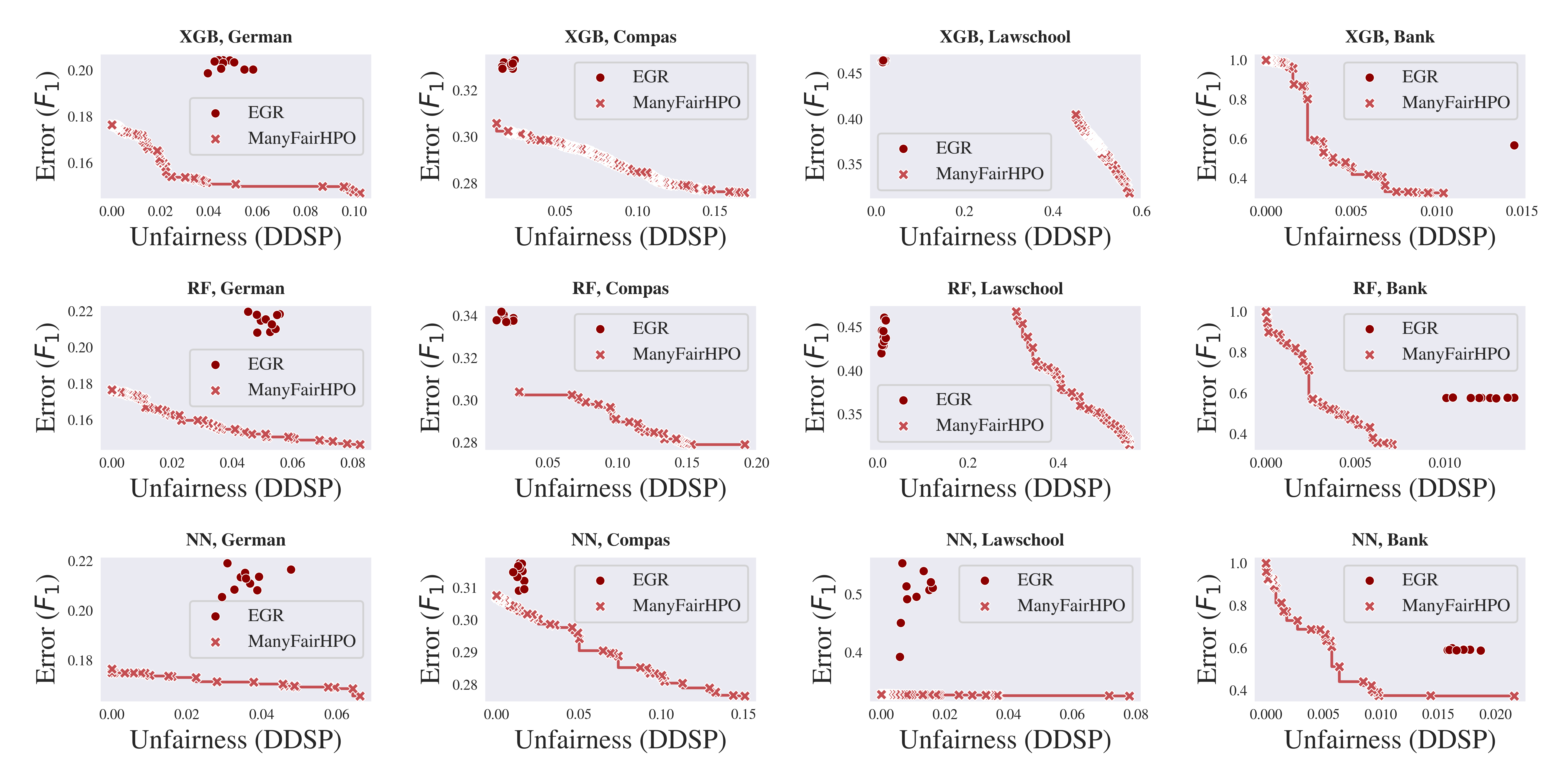}
    \includegraphics[width=\textwidth]{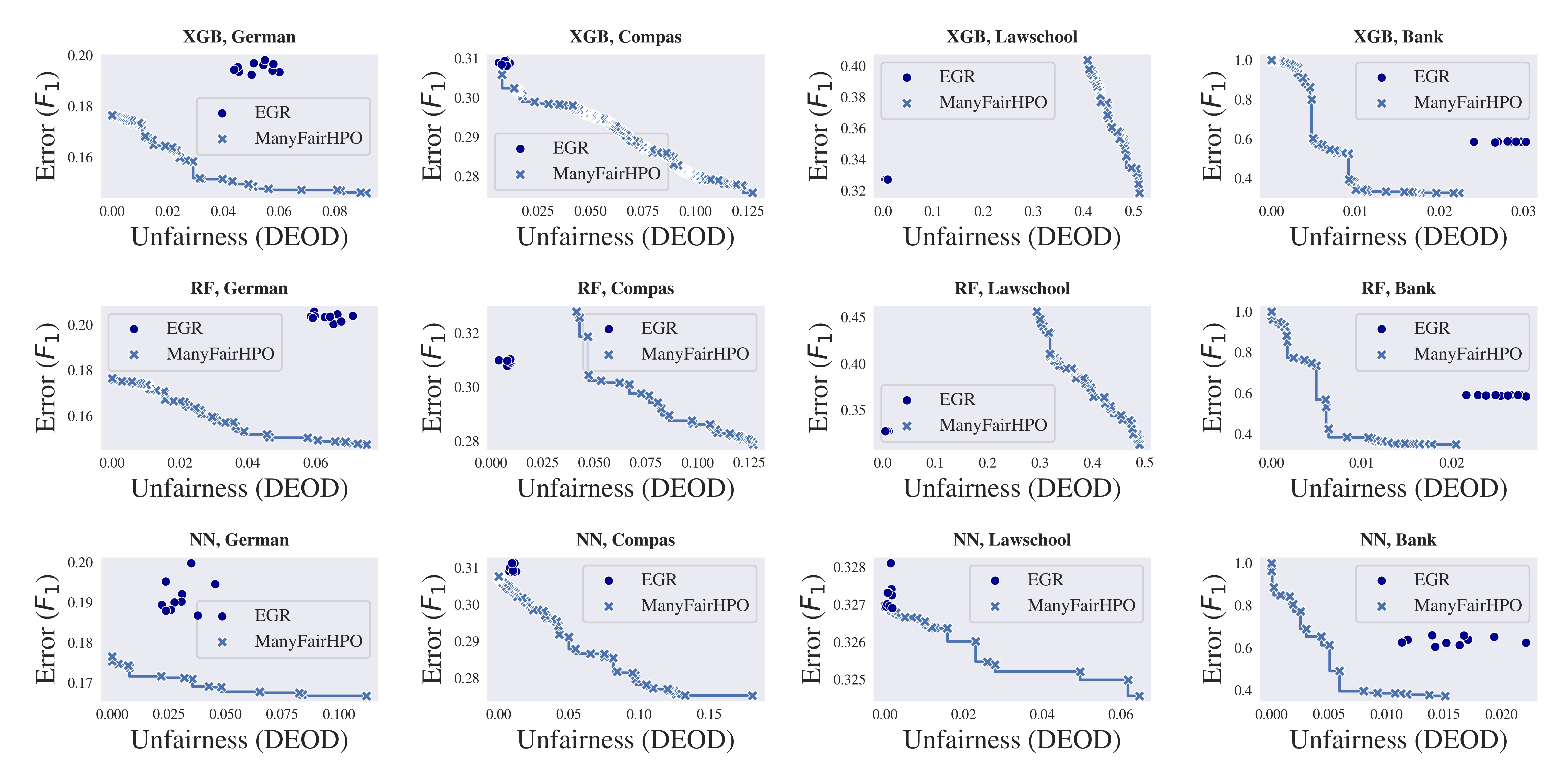}
    \caption{\textit{ManyFairHPO vs. EGR}: Relative fairness-accuracy objective space locations of hyperparameter configurations found by ManyFairHPO and those post-processed with Exponentiated Gradient Reduction (EGR) to minimize DDSP (top) and DEOD (bottom). ManyFairHPO Pareto Fronts dominate EGR models in the majority of cases (9/12 for DDSP and 7/12 for DEOD), suggesting that HPO alone is a competitive approach to bias-mitigation.}
    \label{fig:bias_mit_ddsp}
\end{figure*}

\section{Experimental Details}
\label{section:details}

\subsection{Multi-Criteria Objective Function}

Our objective function takes as input a hyperparameter configuration $\lambda \in \Lambda$, a FairML data set $\mathcal{D} = (X, Y, A)$, and a subset of the fairness metrics $\{f_0,f_1, f_2, ..., f_d\}$.
The objective function applies Nested Stratified k-Fold Cross-Validation to iteratively partition the data set into training, testing, and validation folds $\mathcal{D}_{train}, \mathcal{D}_{val}$ and, $\mathcal{D}_{test}$. Each fold is stratified by both the target $Y$ and protected attribute $A$ in order to maintain a realistic distribution of these variables.

Given a candidate hyperparameter configuration $\lambda \in \Lambda$, a model $\mathcal{M}$ is defined and fit to the training fold $\mathcal{D}_{train}$, generating predictions $\hat{Y}$ on the validation set $\mathcal{D}_{val}$. The predictive performance of the hyperparameter configuration $f_0(Y, \hat{Y})$ is calculated using the $F_1$-Score, an appealing performance metric in the face of significant class imbalance. Because a higher $F_1$-Score is better with respect to predictive performance and defined in the range $(0, 1)$, we minimize $f_0(Y, \hat{Y}):=1-F_1$ during optimization. The \textit{unfairness} of the hyperparameter configuration $f_{1:d}(Y, \hat{Y}, A)$ is calculated using the measures of fairness defined in Table \ref{table:metrics}. The objective values of each evaluated hyperparameter configuration are added to an archive of all observations $\mathcal{Y}$.

\begin{table*}[t!]

\begin{minipage}{\linewidth}
  \centering
    \begin{tabular}{lll}
     \multicolumn{2}{c}{Random Forest (NN)} \\
    \toprule
    Name & Range & Scale \\
    \midrule
     \texttt{max\_depth}  & (1, 50) & Log    \\
     \texttt{min\_samples\_fold}  & (2, 128) & Log     \\
     \texttt{min\_samples\_leaf}  & (1, 20) & Uniform    \\
     \texttt{max\_features}  & (0, 1)  & Uniform   \\
     \texttt{n\_estimators}  & (1, 200) &  Log   \\
     \midrule
  \end{tabular}
   \begin{tabular}{lll}
     \multicolumn{2}{c}{XGBoost (XGB)} \\
    \toprule
    Name & Range & Scale \\
    \midrule
     \texttt{eta}  & ($2^{-10}$, 1.0)  & Log    \\
     \texttt{max\_depth}  & (1, 50)  & Log   \\
     \texttt{colsample\_bytree}  & (0.1, 1.0)  & Uniform   \\
     \texttt{reg\_lambda}  & ($2^{-10}$, $2^{10}$) & Log    \\
     \texttt{n\_estimators}  & (1, 200) & Log    \\
     \midrule
  \end{tabular}
  \begin{tabular}{lll}
  \multicolumn{2}{c}{Multi-Layer Perceptron (NN)} \\
    \toprule
    Name & Range & Scale \\
    \midrule
     \texttt{depth}  & (1, 3) & Uniform   \\
     \texttt{width}  & (16, 1024)  & Log   \\
     \texttt{batch\_size}  & (4, 256) & Log    \\
     \texttt{alpha}  & ($10^{-8}$, 1) & Log    \\
     \texttt{learning\_rate\_init}  & ($10^{-5}$, 1) & Log    \\
     \texttt{n\_iter\_no\_change}  & (1, 20) & Log   \\
  \midrule
  \end{tabular}
\end{minipage}
\\
  \caption{\textit{HPO Search Spaces}: Summary of hyperparameter search spaces drawn from HPOBench.}
  \label{table:hparams}
\end{table*}

\begin{table*}[h!]

\begin{minipage}{\linewidth}
  \centering
    \begin{tabular}{lllcccccc}
    \toprule
Formulation & Name & Optimizer & Objectives & Pop. Size & Func. Evals. & Seeds \\
    \midrule
      BiO & F1-DDSP & NSGA-II & 2 & 20 & 1000 & 10 \\

    & F1-DEOD & NSGA-II & 2  & 20 & 1000 & 10 \\
    & F1-DEOP & NSGA-II & 2  & 20 & 1000 & 10 \\

    & F1-INVD & NSGA-II & 2  & 20 & 1000 & 10 \\
    MaO & F1-MULTI & NSGA-III & 5  & 42 & 1000 & 10 \\
      \midrule
  \end{tabular}
\end{minipage}
\\
  \caption{\textit{ManyFairHPO Experiments:} Summary of ManyFairHPO experiments, spaning across two problem formulations, four fairness metrics, three HPO search spaces, and five data sets. We run each experiment for 10 seeds with a maximum wall-clock time of 1 CPU day.}
  \label{table:experiment}
\end{table*}
\begin{table*}
\begin{minipage}{\linewidth}
  \centering
    \begin{tabular}{lccccc}
    \cmidrule(r){1-6}
    Name & Prot. Attr. & Samples & Features & Pos./Neg. & Priv./Unpriv. \\
    \midrule
     German Credit & sex & 1,000 & 59 & 70/30 & 69/31 \\
     Criminal Recidvism & race & 5278 & 7 & 53/47 & 40/60 \\
     Bank Marketing & age & 764 & 31 & 23/77 & 64/36\\
     Census Income & sex & 15,315 & 44 & 25/75 & 85/14
     \\
     Lawschool Admissions & race & 22,342 & 3 & 25/75 & 92/8  \\
     \midrule
   \end{tabular}
\end{minipage}
\\
  \caption{\textit{FairML Datasets:} Summary of data sets drawn from the \texttt{aif360} library.}
    \label{table:dataset}
\end{table*}

% \begin{figure}
%     \centering
%     \includegraphics[width=0.24\linewidth]{figures/rq2/multi_pareto_rf_adult_ddsp.png}
%     \includegraphics[width=0.24\linewidth]{figures/rq2/multi_pareto_rf_adult_deod.png}
%     \includegraphics[width=0.24\linewidth]{figures/rq2/multi_pareto_rf_adult_deop.png}
%     \includegraphics[width=0.24\linewidth]{figures/rq2/multi_pareto_rf_adult_invd.png}

%     \includegraphics[width=0.24\linewidth]{figures/rq2/multi_pareto_rf_lawschool_ddsp.png}
%     \includegraphics[width=0.24\linewidth]{figures/rq2/multi_pareto_rf_lawschool_deod.png}
%     \includegraphics[width=0.24\linewidth]{figures/rq2/multi_pareto_rf_lawschool_deop.png}
%     \includegraphics[width=0.24\linewidth]{figures/rq2/multi_pareto_rf_lawschool_invd.png}
%     \caption{\textit{MaO Pareto Fronts}: Visualization of the MaO Pareto Front in the presence of a fairness metric conflicts observed on the RF-Adult (top) and RF-Lawschool (bottom) experiments. ManyFairHPO has the consistent effect of filling the gaps left by bi-objective optimization.}
%     \label{fig:conflict_subs}
% \end{figure}

\bibliography{bib/strings,bib/lib,bib/local,bib/proc}